\documentclass{article}

     \PassOptionsToPackage{numbers, compress}{natbib}
\usepackage{graphicx}
\usepackage{wrapfig}
\usepackage[final]{neurips_2023}

\usepackage[utf8]{inputenc} % allow utf-8 input
\usepackage[T1]{fontenc}    % use 8-bit T1 fonts
\usepackage{hyperref}       % hyperlinks
\usepackage{url}            % simple URL typesetting
\usepackage{booktabs}       % professional-quality tables
\usepackage{amsfonts}       % blackboard math symbols
\usepackage{amsmath}
\usepackage{nicefrac}       % compact symbols for 1/2, etc.
\usepackage{microtype}      % microtypography
\usepackage{xcolor}         % colors

\newtheorem{takeaway}{Takeaway}

\newcommand{\norm}[1]{\left \Vert #1 \right \Vert}
\newcommand{\scalarprod}[2]{\left \langle #1, #2 \right \rangle}

\title{Large Learning Rates Improve Generalization: \\ But How Large Are We Talking About?}

\author{Ekaterina Lobacheva$\bf{}^{1}$\thanks{First two authors contributed equally.}\:, Eduard Pockonechnyy$\bf{}^{1}$\footnotemark[1]\:, Maxim Kodryan$\bf{}^{2}$, Dmitry Vetrov$\bf{}^{3}$\\
 ${}^1$Independent researcher \quad  ${}^2$HSE University \quad ${}^3$Constructor University, Bremen \\
	{\tt lobacheva.tjulja@gmail.com, epokonechnyy@gmail.com} \\
 {\tt mkodryan@hse.ru, dvetrov@constructor.university} 
}

\begin{document}

\maketitle

\begin{abstract}
  Inspired by recent research that recommends starting neural networks training with large learning rates (LRs) to achieve the best generalization, we explore this hypothesis in detail.
  Our study clarifies the initial LR ranges that provide optimal results for subsequent training with a small LR or weight averaging.
  We find that these ranges are in fact significantly narrower than generally assumed.
  We conduct our main experiments in a simplified setup that allows precise control of the learning rate hyperparameter and validate our key findings in a more practical setting.
\end{abstract}

\section{Introduction}

In recent years, there has been a strong opinion in the deep learning community about the need to start neural networks training with a large learning rate (LR) to achieve optimal quality.
This is evidenced by both empirical research and theoretical analysis.
Initial large LR values are known to help avoid bad local minima~\cite{lewkowycz2020large,mohtashami2022avoiding} and find better generalizing solutions~\cite{iyer2020wide,kodryan2022training}.
Possible mechanisms for this include increasing the level of implicit regularization in stochastic gradient descent (SGD)~\cite{barrett2021implicit,smith2021origin}, encouraging simplicity of the solutions~\cite{andriushchenko2023sgd,chen2023stochastic}, changing the order of learning patterns in data~\cite{li2019towards},~etc.

However, the existing literature still does not have a clear answer to the question: \emph{how large an LR should we take to achieve the best quality?}
A general recommendation is to start training with an LR value that is too high for convergence but too low for divergence~\cite{andriushchenko2023sgd,iyer2020wide,kodryan2022training}, with no precise benchmarks available over this wide range.
We conduct a detailed empirical analysis dedicated to this problem.

We investigate the properties of the final solutions obtained by training with small learning rates or weight averaging after initial training with different LRs and find that the dependence is, generally speaking, very nontrivial.
We indeed find that the best choice is to start training with LRs that lead to neither convergence nor divergence, but only a relatively narrow portion of that range provides consistent optimal results.

\section{Methodology}
 
To rigorously study the influence of the initial large LR on the final solution, we need to train neural networks with a fixed LR. 
Unfortunately, since most modern architectures use normalization in some form, truly fixing an LR becomes a nontrivial action. 
Specifically, scale-invariance induced by normalization yields two consequences: 1) scale-invariant weights are essentially defined on the sphere, and 2) the \emph{effective learning rate} (ELR) of these weights, i.e., learning rate on the unit sphere, is varying even with a fixed LR due to a varying parameters norm~\cite{arora2019theoretical,li2020reconciling,lobacheva2021periodic,kodryan2022training,nakhodnov2022loss}.
Therefore, following the recent related work of~\citet{kodryan2022training}, we train our models in a fully scale-invariant setting on the unit sphere.
In other words, we make all our models fully scale-invariant by fixing the last layer and removing trainable batch norm parameters, and we train them using projected SGD on the sphere of fixed radius.
This tackles the described issue and thus helps us study optimization with large learning rates in a more accurate fashion, since fixing LR now leads to a fixed ELR as well. 
We return to a more conventional setup in Section~\ref{sec:practice}.

\begin{wrapfigure}{r}{0.37\textwidth}
\vspace{-\baselineskip}
  \centering
  \centerline{
  \includegraphics[width=0.35\textwidth]{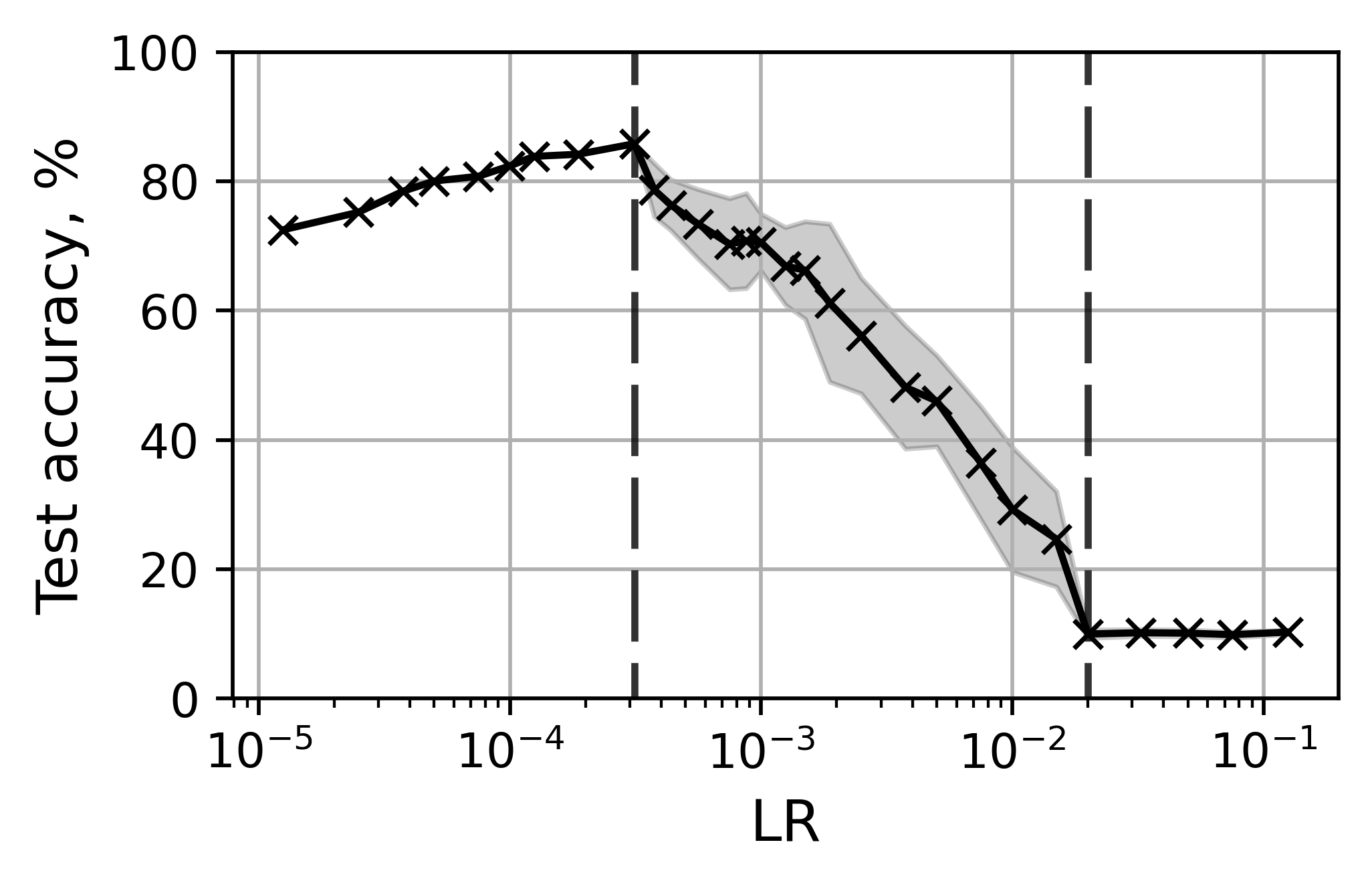}
  }
  \caption{Three regimes in training with fixed LR. Mean test accuracy $\pm$ standard deviation on the last 20 out of 200 epochs are shown. Dashed lines denote boundaries between the training regimes. ResNet-18 on CIFAR-10.}
  \label{fig:three_regimes_demo}
  \vspace{-\baselineskip}
\end{wrapfigure}

Among other things,~\citet{kodryan2022training} discovered that training scale-invariant models on the sphere typically occurs in three regimes depending on the LR value: 1) convergence, when parameters monotonically converge to a minimum, 2) chaotic equilibrium, when loss noisily stabilizes at some level, and 3) divergence, when a model has random guess accuracy.
We adopt the three-regimes taxonomy and build our analysis around it from here on.
In Figure~\ref{fig:three_regimes_demo}, we show the (smoothed) test accuracy after training with different fixed LRs. 
The three regimes are clearly distinguishable and exhibit the expected behavior.
\citet{kodryan2022training} mostly focus on training with fixed LRs, however, they point out that starting training in the second regime and then decreasing LR to any value from the first regime can lead to solutions similar in quality to the ones obtained with optimal first regime LRs. 
Following these results, in this work we examine more closely how initial training with high LRs influences final solutions.
In particular, we are interested in whether starting training in the second regime might be even more beneficial than training with any fixed LR from scratch, in terms of final test accuracy.

We wish to analyze the points obtained after initial training with given LR values from the perspective of their utility for subsequent training with small LRs or weight averaging.
To this end, we divide training into two stages. 
First, we perform so-called \emph{pre-training}, i.e., we train models with different fixed LRs, which we call PLRs, for sufficient amount of epochs to ensure stabilization of training dynamics.
After pre-training, we either (1) change the learning rate and \emph{fine-tune} the model, i.e., train it further with a small LR, or (2) continue training with the same LR as at the pre-training stage and weight-average consequent checkpoints, as is usually done in stochastic weight averaging (SWA)~\cite{izmailov2018averaging}.
For fine-tuning, we use only first regime LRs, which we call FLRs, to ensure model convergence.

We compare the obtained solutions in terms of their generalization and geometry.
In particular, we evaluate the test accuracy of the fine-tuned/SWA models and analyze the linear connectivity and angular distance between them.
Angular distance between two networks with weights $\theta_1$ and $\theta_2$ is calculated as 
\[angle(\theta_1, \theta_2) = \arccos\left(\frac{\scalarprod{\theta_1}{\theta_2}}{\norm{\theta_1} \norm{\theta_2}}\right). \]
We choose it as a natural metric on the sphere in the weight space. 
Linear connectivity is measured via calculating a linear-path barrier between two networks w.r.t. training or test error, i.e., the highest difference between the error on the linear path between two points in the weight space and linear interpolation of the error at each of them~\cite{entezari2022role}:
\[B(\theta_1, \theta_2) = \sup_{\alpha \in [0, 1]} \left[ L(\alpha \theta_1 + (1 - \alpha) \theta_2) - \alpha L(\theta_1) - (1 - \alpha) L(\theta_2) \right], \]
where $\theta_1$ and $\theta_2$ are weights of the networks, and $L$ is the error measure.
The barrier value shows whether solutions obtained from the same pre-trained point remain in the same low-error region (low barrier) or head to different optima (high barrier).
Further detail on the setup can be found in Appendix~\ref{app:setup}.

\section{Main results}

\begin{figure}
  \centering
  \centerline{
  \includegraphics[width=\textwidth]{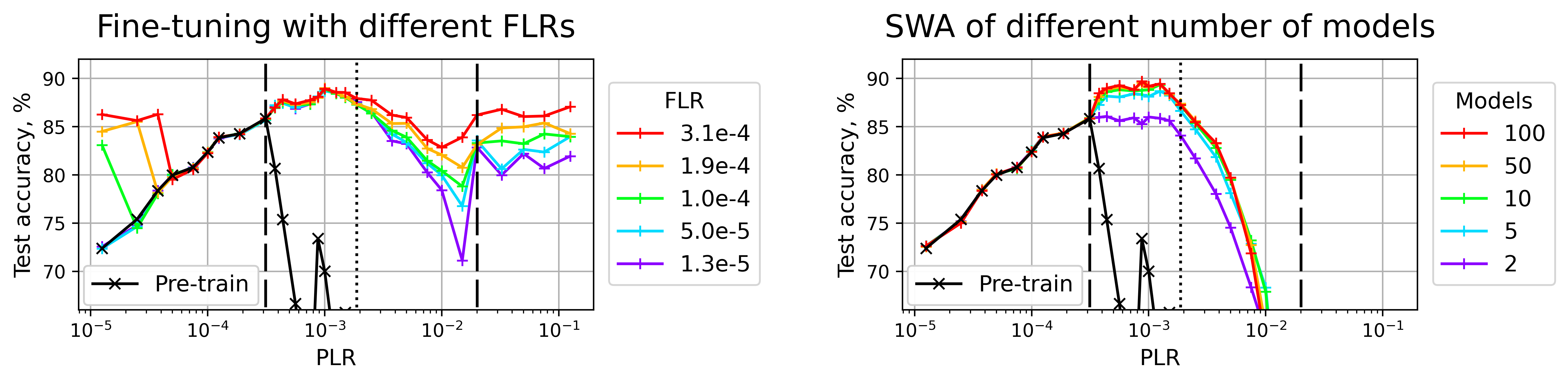}
  }
  \caption{Test accuracy of different fine-tuned (left) and SWA (right) solutions. Test accuracy after pre-training is depicted with the black line. Dashed lines denote boundaries between the pre-training regimes, dotted line divides the second regime into two subregimes.}
  \label{fig:accuracy}
\end{figure}

\begin{figure}
  \centering
  \centerline{
  \includegraphics[width=\textwidth]{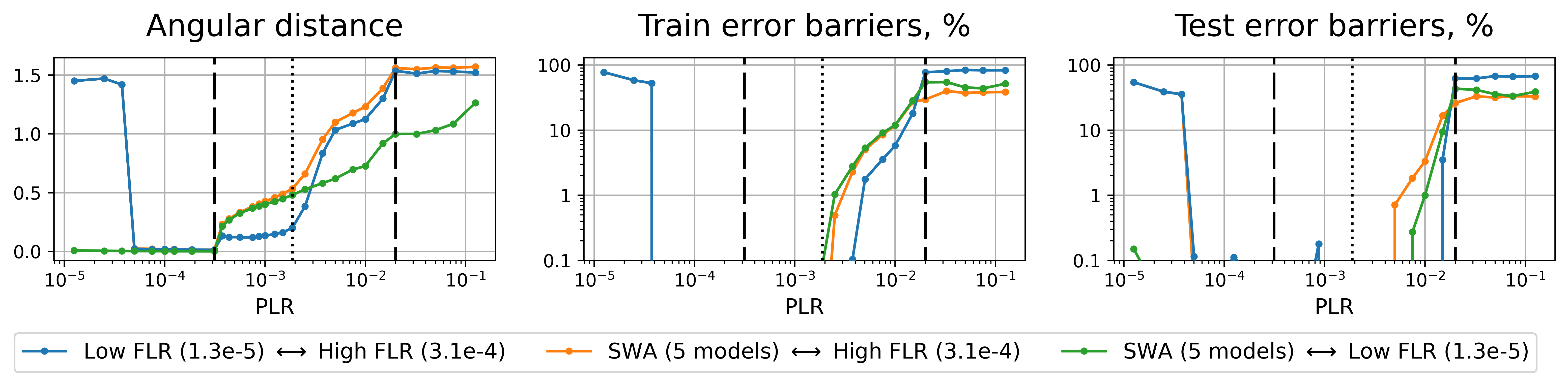}}
  \caption{Geometry between the points fine-tuned with the smallest and the largest FLRs and SWA.}
  \label{fig:analysis}
\end{figure}

In this section, we provide our main results obtained with a scale-invariant ResNet-18~\cite{deep_resnet} trained on CIFAR-10~\cite{krizhevsky2009learning}.
We also consider a ConvNet architecture and CIFAR-100 dataset in Appendix~\ref{app:dataset_architecture}.
Figure~\ref{fig:accuracy} depicts the test accuracy after pre-training with different PLR values (black line) and after fine-tuning with given FLRs or SWA of given numbers of checkpoints (colored lines).
In Figure~\ref{fig:analysis}, we also present angular distances and linear connectivity between three solutions for each PLR: SWA of 5 networks and the points obtained after fine-tuning with the lowest and the highest considered FLRs.
We divide the pictures into three parts (using dashed lines) corresponding to the three previously introduced pre-training regimes and discuss them separately in the following subsections.

\subsection{Pre-training in regime 1}
Training in the first training regime typically leads to convergence to a minimum.
In accordance with the results of~\citet{kodryan2022training} we observe a monotonic dependence between the PLR value and the test accuracy of the pre-trained model. 
Fine-tuning with FLR $\leq$ PLR and SWA cannot significantly improve the pre-trained model because they remain very close to it.
Hence, they result in solutions of similar quality, which are close to each other in angular distance and can be linearly connected.
At the same time, FLR $>$ PLR may lead to training instabilities and subsequent convergence to a new better minimum~\cite{lewkowycz2020large}.
This effect can be clearly observed when fine-tuning with the highest FLR from low PLRs through improvement of test accuracy in Figure~\ref{fig:accuracy} and high angular distance and error barrier with other solutions in Figure~\ref{fig:analysis}.
However, the training instabilities only happen when FLR is substantially larger than PLR.
Therefore, the best strategy of training in the first regime is to use the maximum constant LR without any schedules.

\begin{takeaway}
\label{take:regime_1}
Even though pre-training with the largest first regime PLR gives a 
model with the best test accuracy, it cannot be improved via fine-tuning or weight averaging. 
Hence, LRs of the first regime are not the best choice to use at the beginning of training. 
\end{takeaway}

\subsection{Pre-training in regime 2}

The second regime can be considered the most important for pre-training, since most practical LR schedules start from this regime~\cite{kodryan2022training}.
We can see that the results strongly depend on the PLR value and the general advice ``pre-train with a large LR to obtain a better solution'' is valid, but not for all PLRs of the second regime.
In fact, the second regime can be divided into two subregimes, which we call 2A and 2B.
We discovered that pre-training in subregime 2A, i.e., with lower second regime PLRs, results in significantly better fine-tuning and SWA results compared to the first regime, while pre-training in subregime 2B, i.e., with higher second regime PLRs, loses this advantage and leads to poor fine-tuning and SWA quality.

\paragraph{Pre-training in subregime 2A}
Even though pre-training with lower second regime PLRs does not converge to the low loss region, it locates optimal regions for further fine-tuning or weight averaging.
Fine-tuning a network obtained with a PLR from this range with any FLR results in effectively the same minimum: the obtained points have the same quality, are close to each other in angular distance and are linearly connected.
So, such a pre-training allows fine-tuning even with small FLRs to avoid local optima with poor generalization, to which training from scratch with the same FLRs usually converges.
Moreover, the obtained fine-tuned models are of higher quality than any solutions obtained in the first regime, which shows that the best minima may be completely unreachable with small LRs from a standard initialization, at least in any reasonable training time.
We discuss this hypothesis in more detail in Appendix~\ref{app:1_2_boundary}.
SWA also results in high-quality models, located farther from the fine-tuned solutions but still linearly connected to them. 
Hence, we conjecture that pre-training in subregime 2A locates a ``bowl'' in the loss landscape with a region of high-quality solutions in the middle of it, and further fine-tuning or weight averaging persistently find these solutions. 
Interestingly, SWA shows almost the same quality for all PLRs in this range, while the dependence of fine-tuning test accuracy on PLR is non-monotonic and has a clear maximum.
Explaining this effect is an intriguing challenge and we leave it for future research.

\paragraph{Pre-training in subregime 2B}
Pre-training with higher second regime PLRs gets stuck at higher loss levels and loses the ability to locate the ``bowl'' of high-quality solutions. 
SWA still improves the accuracy at these PLRs but shows worse quality compared to SWA at lower PLRs.
Fine-tuning is more effective than SWA but it results in worse solutions than training from scratch with the same FLRs. 
So, such pre-training can be detrimental if we decrease LR right after it. 
However, this effect can be mitigated by gradually decreasing an LR through the lower second regime LR values (see Appendix~\ref{app:2_end}).
With larger PLRs, fine-tuned solutions not only become worse but also start to differ from each other: we see a rapidly growing gap in test accuracy and angular distance between the solutions obtained with low and high FLRs, followed by a separation in the linear connectivity w.r.t. the training error.
Interestingly, linear connectivity w.r.t. the test error is still mostly preserved.
Hence, we conclude that pre-training in subregime 2B
locates a vast area of the loss landscape with a diverse set of minima.
Obtaining high-quality results in this area with a constant LR or SWA is not possible and requires a more complex LR schedule.

\begin{takeaway}
\label{take:regime_2}
    Pre-training with a lower second regime PLRs helps robustly increase the quality to a level that cannot be achieved in the first regime, however pre-training with higher second regime PLRs loses this advantage and, in contrast, degrades performance.
\end{takeaway}

\subsection{Pre-training in regime 3}

Pre-training in the third regime is somewhat similar to random walking in the parameters space~\cite{kodryan2022training,nakhodnov2022loss}.
Despite this, we notice that it can be beneficial for fine-tuning with a small FLR.
We conjecture that the point obtained by pre-training in regime 3, in contrast to the random initialization, has a very uneven distribution of norms of individual scale-invariant parameter groups: many groups have a low norm, which means that the effective learning rate for them is actually higher than the total ELR for the whole model, which promotes convergence to overall 
better optima (see Appendix~\ref{app:3_regime}).

\begin{takeaway}
\label{take:regime_3}
    Although pre-training in regime 3 does not seem to extract any useful information from the data, it can help in subsequent fine-tuning by providing better initialization for optimization.
\end{takeaway}

\section{Practical setting}

\label{sec:practice}

\begin{figure}
  \centering
  \centerline{
  \includegraphics[width=\textwidth]{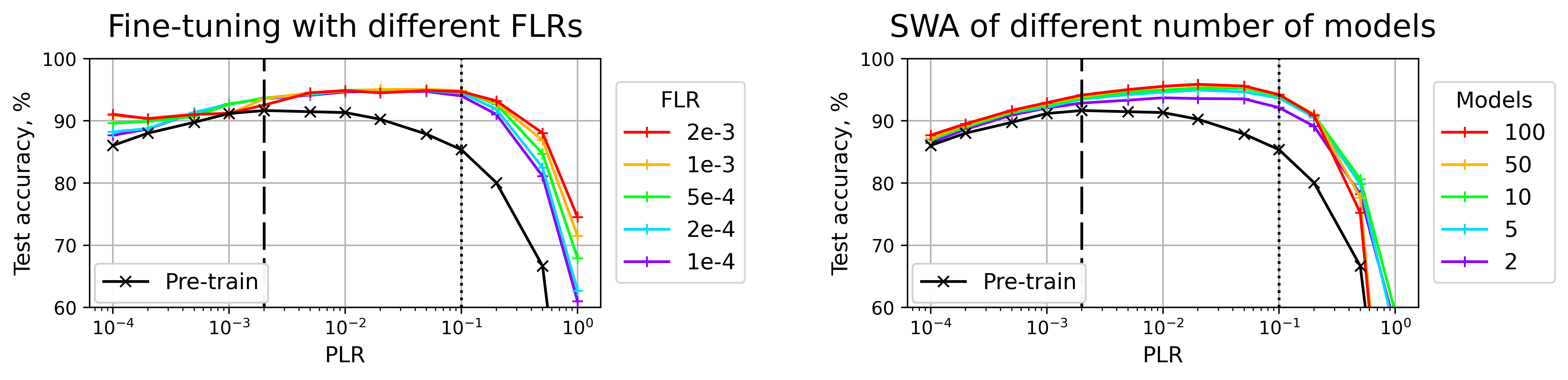}
  }
  \caption{Practical setting. Test accuracy of different fine-tuned (left) and SWA (right) solutions. Test accuracy after pre-training is depicted with the black line. Dashed lines denote boundaries between the first and second pre-training regimes, dotted line divides the second regime into two subregimes.}
  \label{fig:practice}
\end{figure}

In this section, we demonstrate that most of our results can be transferred to a more conventional training setting.
We train a common ResNet-18 model in the whole weight space (without the sphere constraint) using SGD with momentum, weight decay, and data augmentation; the only deviation from the standard setup is a different LR schedule.
Due to unstable behavior of non-scale-invariant weights with large LRs~\cite{kodryan2022training}, we can only observe the first two regimes.
The boundary between regimes 1 and 2 is also less clear, mainly due to the presence of augmentations, which make the data more difficult to learn.
Therefore, the model does not converge completely with low LRs in the first regime, so we have drawn the boundary between the regimes approximately according to the PLR with maximum quality at the pre-trained point.

As can be seen in Figure~\ref{fig:practice}, our Takeaway~\ref{take:regime_2} that the best quality is achieved with LR drops and SWA at the beginning of the second regime is confirmed.
The test accuracy optimum is also achieved not at the very beginning, but a little further into the second regime.
It is also clear that fine-tuning and SWA at the beginning of the second regime converge to similar optima, while at large PLRs fine-tuning leads to diverse minima and SWA quality rapidly deteriorates.
In the first regime, the behavior is substantially different due to the mentioned issues with convergence: we see that both SWA and fine-tuning even with lower LRs can still improve the quality. 
Interestingly, both fine-tuning and SWA at the best points achieve a quality close to the conventional $\sim95\%$ test accuracy for ResNet-18 on CIFAR-10 trained with a standard LR schedule.

\begin{takeaway}
\label{take:practice}
    The majority of the described properties of the final solutions obtained after pre-training with different LR values remain relevant for practical applications.
\end{takeaway}

\section{Conclusion}

In this work, we studied the influence of pre-training a neural network with different LRs on the quality of the final solution.
We discovered that initial training with moderately large LRs, just short of convergence, provides the best opportunity for subsequent fine-tuning or weight averaging.
Our findings pave the way for important future work on optimizing the strategy for controlling the learning rate and also shed more light on the overall loss landscape structure.

\begin{ack}
We would like to thank Ildus Sadrtdinov and the anonymous
reviewers for their valuable comments.
Maxim Kodryan was supported by the grant for research centers in the field of AI provided by the Analytical Center for the Government of the Russian Federation (ACRF) in accordance with the agreement on the provision of subsidies (identifier of the agreement 000000D730321P5Q0002) and the agreement with HSE University \textnumero 70-2021-00139.
The empirical results were supported in part through the computational resources of HPC facilities at HSE University~\citep{hpc}.
\end{ack}

\bibliographystyle{apalike}
\bibliography{ref}

\newpage
\appendix

\section{Experimental setup}
\label{app:setup}

{\bf Datasets and architectures.} 
Following~\citet{kodryan2022training}, we conduct experiments with two network architectures, a simple 3-layer convolutional neural network with batch norm layers (ConvNet) and a ResNet-18, on CIFAR-10 and CIFAR-100 datasets. 
In scale-invariant setup, we use ResNet models with width factor $k=32$, and ConvNet models with width factor $k=32$ and $128$ for CIFAR-10 and CIFAR-100 respectively. 
In practical setup, we use ResNet with a standard width factor $k=64$.

{\bf Pre-training, fine-tuning and SWA.} 
We train all networks using stochastic gradient descent with a batch size of 128. 
Both the pre-training and the fine-tuning stages take $200$ epochs.
This training time is sufficient to either reach a minimum or stabilize the loss in the pre-training stage and to achieve complete convergence in the fine-tuning stage even with the smallest FLRs.  
Fine-tuning is always done with the first regime FLRs to ensure convergence to a minimum.
When performing SWA of $N$ models, we continue training for $N-1$ more epochs with the same PLR and average checkpoints from epochs $200, \dots, 200 + N - 1$.
In the practical setting in Section~\ref{sec:practice}, we use weight decay of $5 \cdot 10^{-4}$, momentum of $0.9$, and standard CIFAR augmentations: random crops (size: $32$, padding: $4$) and random horizontal flips. 

\section{General results on other datasets and architectures}
\label{app:dataset_architecture}

\begin{figure}
  \centering
  \centerline{
  \begin{tabular}{c}
  ConvNet on CIFAR-10\\
  \includegraphics[width=\textwidth]{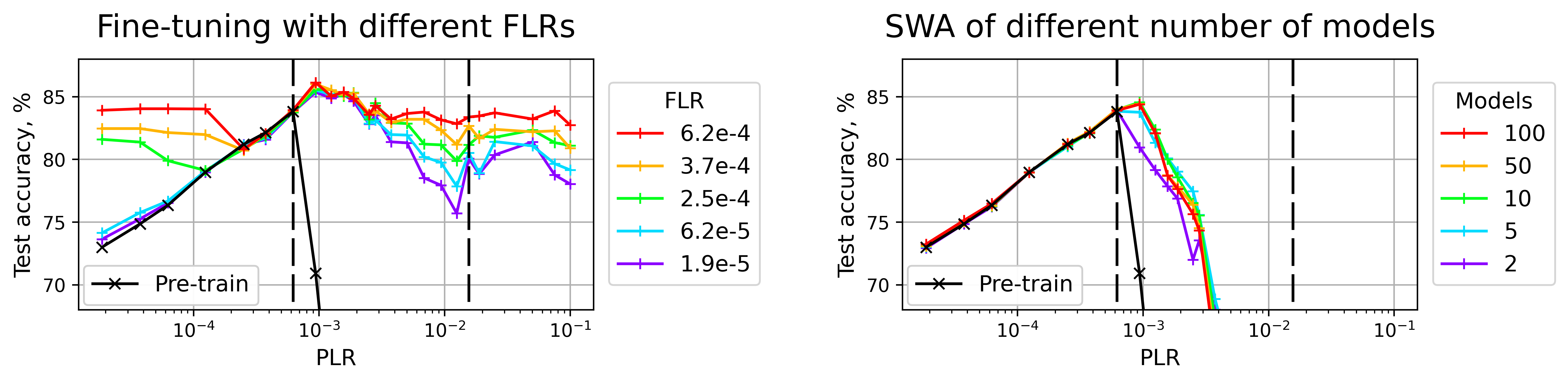}\\
  ConvNet on CIFAR-100\\
  \includegraphics[width=\textwidth]{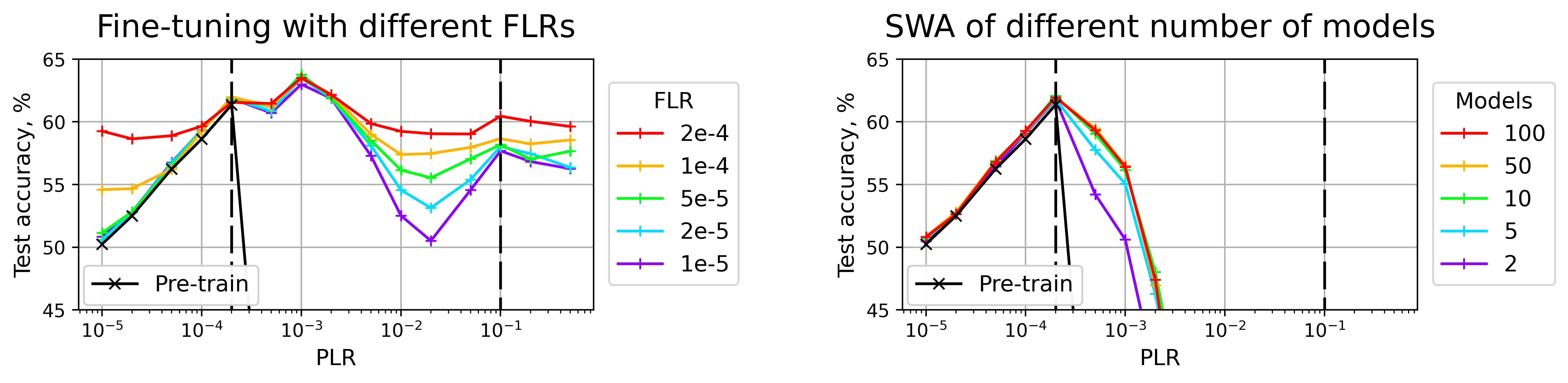}\\
  ResNet-18 on CIFAR-100\\
  \includegraphics[width=\textwidth]{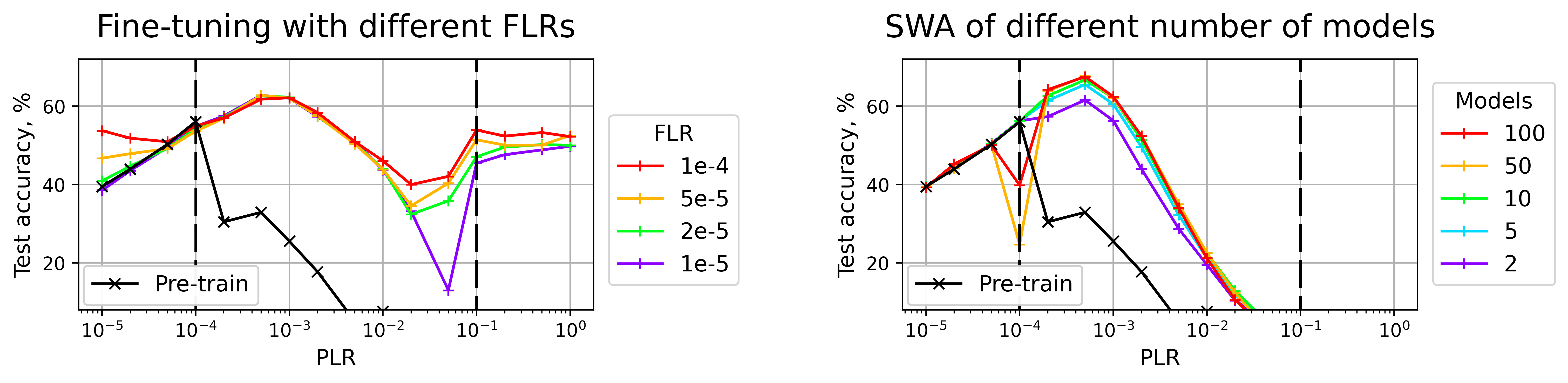}\\
  \end{tabular}
  }
  \caption{Test accuracy of different fine-tuned (left) and SWA (right) solutions. Dashed lines denote boundaries between the pre-training regimes. Results for other dataset-architecture pairs, similar to Figure~\ref{fig:accuracy}.}
  \label{fig:app_accuracy}
\end{figure}

In Figure~\ref{fig:app_accuracy}, we demonstrate the test accuracy after pre-training with different PLR values (black line) and after fine-tuning with given FLRs or SWA (colored lines) for other dataset-architecture pairs: ConvNet on CIFAR-10/CIFAR-100 and ResNet-18 on CIFAR-100.
The results are similar to that of ResNet-18 on CIFAR-10, described in the main text and shown in Figure~\ref{fig:accuracy}.

We again can divide each plot into three parts w.r.t. the three pre-training regimes, and the main takeaways still hold.
Pre-training test accuracy is monotonic in the first regime and both fine-tuning and SWA are unable to improve it for high PLRs of the first regime.
Optimal PLRs for further fine-tuning/SWA are attributed to the beginning of the second regime, while the quality deteriorates by the end of the second regime.
Fine-tuning with low FLRs in the third regime is better than training with the corresponding LR from scratch starting at the standard random initialization.

There are, however, two remarks.
First, for the ConvNet model the effects of the second regime are much less pronounced, i.e., both fine-tuning and SWA show less improvement for low PLRs of the second regime and the deteriorating effect of large PLRs on fine-tuning is almost leveled out.
We suppose that it could be explained by the simplicity of ConvNet, which allows more robust training with a fixed LR and much less scope for further quality improvement. 
Second, due to the periodic behavior~\cite{lobacheva2021periodic} of ResNet-18 trained on CIFAR-100 with high first regime LRs, also reported by~\citet{kodryan2022training}, we observe instabilities for SWA of more than 10 models, since checkpoints from different periods lay in different low-loss regions in that case.

\section{Boundary between the first and second regimes}
\label{app:1_2_boundary}

\begin{figure}
  \centering
  \centerline{
  \includegraphics[width=0.9\textwidth]{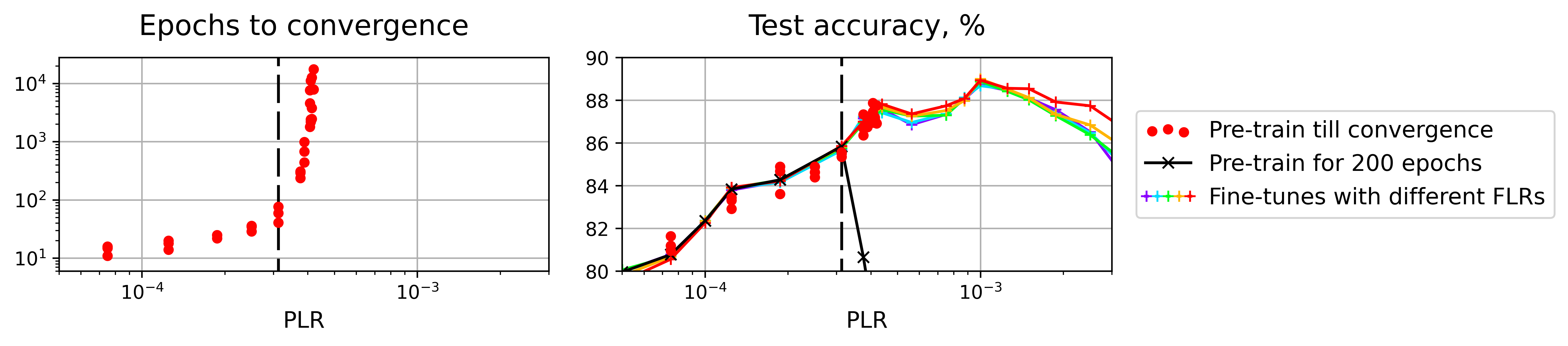}}
  \caption{Number of training epochs to convergence (left) and test accuracy (right) for different PLRs on the boundary between the first and second regimes. Red points denote training to convergence from scratch with a fixed LR value (we run the experiment with three different seeds).}
  \label{fig:1_2_boundary}
\end{figure}

In this section, we motivate our statement in the main text that the best quality obtained after fine-tuning from the second regime is is unattainable by training from scratch with a fixed LR for any reasonable time.
For that purpose, we train our models till convergence (with a maximum of $2 \cdot 10^4$ epochs), i.e., when training loss value reaches $10^{-3}$, and track the convergence time and the achieved test accuracy for different LRs near the boundary between the first and second regimes (see Figure~\ref{fig:1_2_boundary}).

We observe that the required training time increases sharply after the threshold separating the regimes for a $200$ epochs budget. 
The obtained minima have approximately the same test accuracy as the fine-tuned solutions at the corresponding PLRs, however, they take immensely more epochs to reach.
Based on the training time growth, we hypothesize that the best fine-tuning test accuracy obtained at $\text{PLR}=10^{-3}$ is unattainable with any realistic epoch budget.
Therefore, we conclude that training from scratch with a fixed LR from a standard initialization does not allow to find as good optima as become available after pre-training with moderate LRs of the second regime.

\section{Gradual fine-tuning restores the quality for higher second regime PLRs}
\label{app:2_end}

\begin{figure}
  \centering
  \centerline{
  \includegraphics[width=0.55\textwidth]{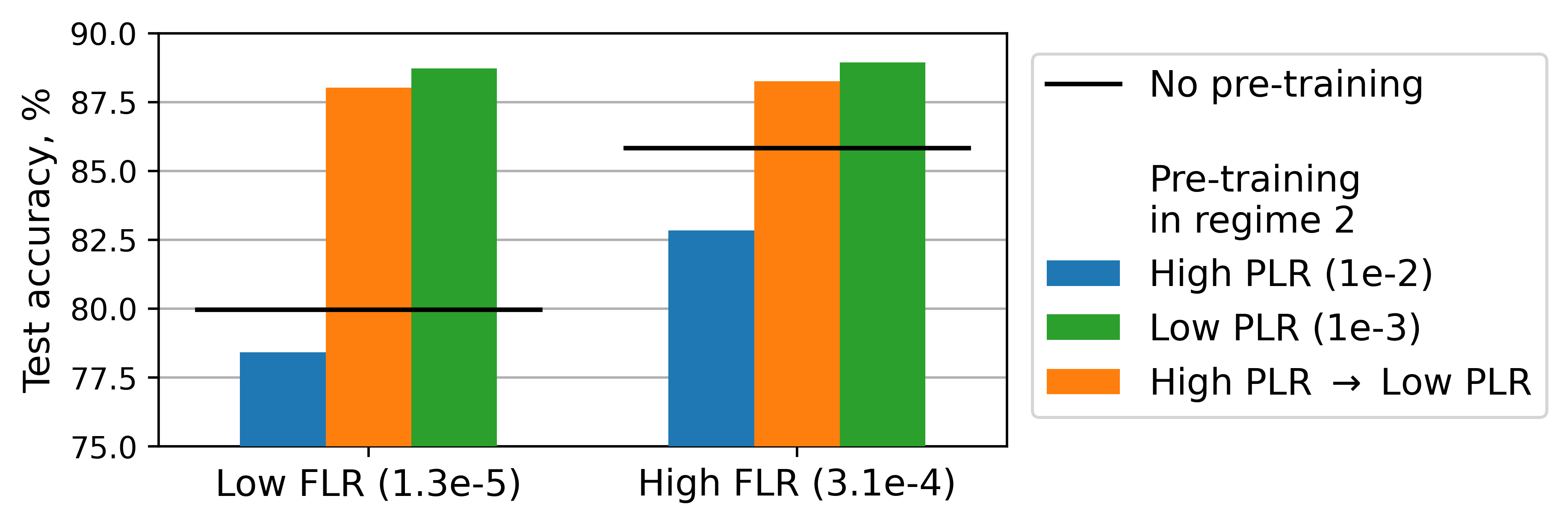}}
  \caption{Test accuracy obtained after fine-tuning with two different FLR values. Blue bar denotes fine-tuning after pre-training with a higher second regime PLR, green bar denotes fine-tuning after pre-training with a lower second regime PLR, and orange bar denotes first fine-tuning with a lower second regime PLR and then with a given FLR after pre-training with a higher second regime PLR. Black lines denote training from scratch with a given FLR.}
  \label{fig:two_pretrainings}
\end{figure}

In this section, we provide additional results on fine-tuning from higher second regime PLRs, attributed to subregime 2B.
As stated in the main text, immediate LR drop leads to worse fine-tuning quality than even training from scratch with the same FLR.
However, we found that adding only one additional step in the LR schedule improves the fine-tuning test accuracy almost to the optimal level. 
Specifically, after pre-training with a PLR of subregime 2B, we first drop LR to a lower PLR of subregime 2A, fine-tune for $200$ epochs, then drop LR once again to a given FLR of the first regime and fine-tune for $200$ more epochs until convergence.
Such a two-step LR schedule helps achieve almost the same test accuracy as usual fine-tuning with the same FLR after pre-training in subregime 2A, which gives the best solution.

Figure~\ref{fig:two_pretrainings} shows test accuracies of the fine-tuned solutions obtained after pre-training with a subregime 2B PLR (blue bars), pre-training with a subregime 2A PLR (green bars), and two-step LR schedule through the subregime 2A PLR (orange bar) for the highest and the lowest FLR values.
Black lines denote the respective test accuracies after training from scratch with the given FLR values.
We can see that indeed gradual fine-tuning through lower second regime LRs restores the quality for higher second regime PLRs.

\section{Pre-training in regime 3: fine-tuning with low FLRs}
\label{app:3_regime}

\begin{figure}
  \centering
  \centerline{
  \includegraphics[width=\textwidth]{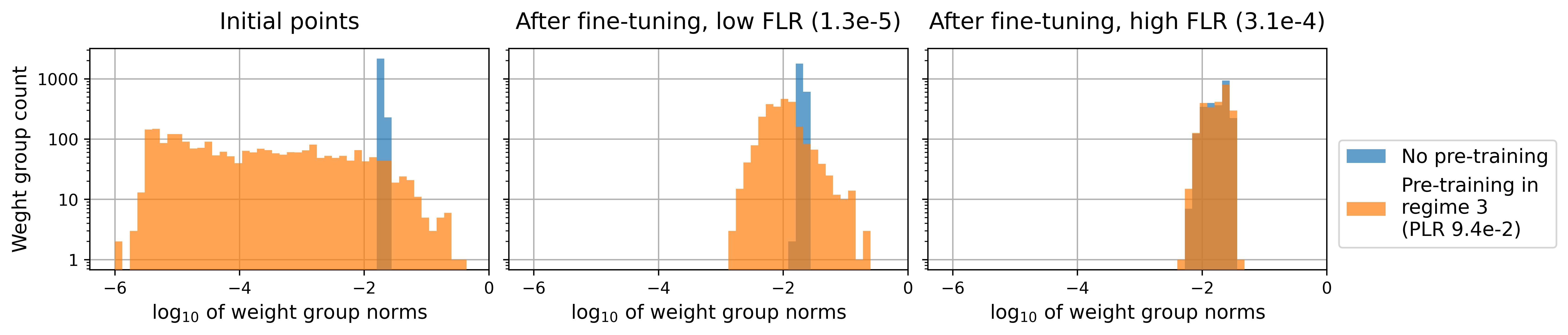}}
  \caption{Histograms of individual scale-invariant weight group norms for standard random initialization (blue) and pre-training with a third regime PLR (orange). Left plot shows norms right after initialization/pre-training, middle plot shows norms after fine-tuning with a low FLR, right plot shows norms after fine-tuning with a high FLR.}
  \label{fig:3_regime_hist}
\end{figure}

In this section, we give more detail on reasons for better fine-tuning results with low FLRs after pre-training in the third regime compared to training from scratch with the same FLRs.
When training with a fixed learning rate $\eta$, ELR for a scale-invariant parameter group $\theta$ is defined as $\eta / \norm{\theta}^2$.
In the main text, we suggest that a possible explanation could be the uneven distribution of norms of individual scale-invariant groups resulting in the corresponding uneven distribution of individual ELRs after pre-training with very large PLRs.
That means, that when fine-tuning from the third regime, some weight groups are learning faster than the others, which gives them the benefits of large learning rate training despite the low total LR.
In contrast, with a standard initialization all weight norms are approximately the same, which implies that the whole model is essentially trained with the same small effective learning rate, resulting in inferior quality. 

In Figure~\ref{fig:3_regime_hist}, we show the histograms of norms of individual scale-invariant weight groups for a standard initialization (blue) and pre-training with a third regime PLR (orange).
We depict three stages: right after initialization/pre-training (left), after training from scratch/fine-tuning with a low FLR (middle), and after training from scratch/fine-tuning with a high FLR (right).
We see that the distribution of weight norms is significantly different with a much more spread distribution after pre-training.
Interestingly, this effect persists after fine-tuning with a low FLR and almost eliminated after fine-tuning with a high FLR, which does not have any advantages over training from scratch.

\end{document}